\ifdefined\pdfoutput\pdfoutput=1\fi
\documentclass[11pt]{article}
\usepackage{amsmath}
\usepackage{enumitem}
\usepackage{fontspec}   
\usepackage{polyglossia}   
\usepackage[final]{acl}   
\usepackage{times}
\usepackage{latexsym}
\usepackage{microtype}
\usepackage{times}
\usepackage{latexsym}
\usepackage[T1]{fontenc}
\usepackage[utf8]{inputenc}
\usepackage{microtype}
\usepackage{inconsolata}
\usepackage{graphicx}
\usepackage[utf8]{inputenc}
\usepackage[table]{xcolor}
\usepackage{subcaption}
\usepackage{pifont}
\usepackage{graphicx}
\usepackage{booktabs}
\usepackage{subcaption}
\usepackage{newunicodechar}
\usepackage{graphicx}
\usepackage{float}
\newunicodechar{✓}{\ding{51}}
\newunicodechar{✗}{\ding{55}}
\setmainlanguage{english} 
\setotherlanguage{arabic} 

\newfontfamily\arabicfont[
  Script=Arabic,
  Path=fonts/,
  Extension=.ttf,
  UprightFont=*-Regular,
  ItalicFont=*-Italic,
  BoldFont=*-Bold,
  BoldItalicFont=*-BoldItalic
]{Amiri}

\usepackage{multirow}

\title{\textsc{AutoArabic}: A Three-Stage Framework for Localizing Video-Text Retrieval Benchmarks\thanks{Accepted at ArabicNLP 2025 (EMNLP 2025 workshop).}}

\author{%
\makebox[\textwidth][c]{%
  \begin{tabular}{@{}c c c@{}}
  Mohamed Eltahir$^{1}$ &
  Osamah Sarraj$^{1}$ &
  Abdulrahman Alfrihidi$^{1}$ \\
  Taha Alshatiri$^{1}$ &
  Mohammed Khurd$^{1}$ & Mohammed Bremoo$^{1}$ \\
  & Tanveer Hussain$^{2}$ 
  \end{tabular}%
}%
\\[0.45em]
$^{1}$ King Abdullah University of Science and Technology (KAUST), Thuwal, Saudi Arabia \\
$^{2}$ Department of Computer Science, Edge Hill University, Ormskirk, England \\
\texttt{\{mohamed.hamid@kaust.edu.sa, osamah.sarraj@gmail.com, frihidimany@gmail.com,} \\
\texttt{tahaalshatiri@gmail.com, mohamedalawi211@gmail.com, mohabremoo@gmail.com} \\
\texttt{hussaint@edgehill.ac.uk\}}
}

\begin{document}
\maketitle



\begin{abstract}
Video-to-text and text-to-video retrieval are dominated by English benchmarks (e.g. DiDeMo, MSR-VTT) and recent multilingual corpora (e.g. RUDDER), yet Arabic remains underserved, lacking localized evaluation metrics. We introduce a three-stage framework, \textsc{AutoArabic}, utilizing state-of-the-art large language models (LLMs) to translate non-Arabic benchmarks into Modern Standard Arabic, reducing the manual revision required by nearly fourfold. The framework incorporates an error detection module that automatically flags potential translation errors with 97\% accuracy. Applying the framework to DiDeMo, a video retrieval benchmark produces DiDeMo-AR, an Arabic variant with 40,144 fluent Arabic descriptions. An analysis of the translation errors is provided and organized into an insightful taxonomy to guide future Arabic localization efforts. We train a CLIP-style baseline with identical hyperparameters on the Arabic and English variants of the benchmark, finding a moderate performance gap ($\Delta \approx 3$\,pp at Recall@1), indicating that Arabic localization preserves benchmark difficulty. We evaluate three post-editing budgets (zero/ flagged-only/ full) and find that performance improves \emph{monotonically} with more post-editing, while the raw LLM output (zero-budget) remains \emph{usable}. To ensure reproducibility to other languages, we made the code available at \url{https://github.com/Tahaalshatiri/AutoArabic}.
\end{abstract}


\begin{table*}[t]
\centering
\small
\caption{\textbf{Video-text retrieval benchmarks}. This table highlights a \textit{language gap}: existing retrieval benchmarks are almost entirely English (with limited Chinese) and lack Arabic coverage. To our knowledge, only our DiDeMo-AR offers Modern Standard Arabic captions. "Moment-level" ✓ indicates that the dataset provides temporally-localized descriptions (segment boundaries).}
\label{tab:retrieval_benchmarks}
\begin{tabular}{@{}lcccccc@{}}
\hline
\textbf{Dataset} & \textbf{\#Videos} & \textbf{Clip Len.} &  \textbf{Languages} & \textbf{Moment-level} & \textbf{Arabic?} \\ \hline
MSR-VTT \cite{xu2016msr}               & 10,000  & 15s   &  EN                & ✗ & ✗ \\
VATEX \cite{wang2019vatex}                & 41,250  & 10s   &  EN / ZH           & ✗ & ✗ \\
DiDeMo \cite{anne2017localizing}                & 10,464  & 30s    & EN                & ✓ & ✗ \\
LSMDC \cite{rohrbach2015dataset}                 & 118,081 & 4-5s     & EN                & ✗ & ✗ \\
ActivityNet \cite{caba2015activitynet}  & 19,994  & 120s  &  EN                & ✓ & ✗ \\
RUDDER \cite{dabral2021rudder}                & 100 k / lang. & 5-10s & EN / ZH / FR / DE / RU & ✗ & ✗ \\ 
\rowcolor{gray!15}
\textbf{DiDeMo-AR}  & 10,464 & 30s &  \textbf{AR} & ✓ & \textbf{✓} \\ \hline
\end{tabular}
\end{table*}

\begin{table*}[t]
\centering
\small
\caption{\textbf{Arabic corpora with different modalities} (non-retrieval). This highlights a \textit{task gap}: prior corpora focus on speech, sentiment, or QA and do not provide video–text retrieval benchmarks. DiDeMo-AR is the first publicly released Arabic dataset dedicated to retrieval.}  
\label{tab:arabic_multimodal}
\begin{tabular}{lcccc}
\toprule
\textbf{Dataset} & \textbf{Modality} & \textbf{Primary Task} & \textbf{Size / Hours} & \textbf{Retrieval?} \\
\midrule
Amd’SaEr \cite{haouhat2023towards}      & Video + Audio + Text & Multimodal Sentiment & 540 clips & ✗ \\
MGB-2 \cite{Ali2016MGB2}                 & Audio + Subtitles    & ASR (broadcast MSA)  & $\sim$1200 h & ✗ \\
MASC \cite{al2023masc}                   & Audio                & ASR (speech corpus)  & $\sim$1000 h & ✗ \\
GALE Arabic \cite{LDC2017T12}              & Audio + Text         & ASR/MT (news/talk)   & multi-year & ✗ \\
ArabicaQA \cite{abdallah2024arabicaqa}   & Text                 & QA / Dense Retrieval & 92k Q/A & ✗ \\
ANAD \cite{elnagar2020anad}              & Audio                & Speech Emotion Rec.  & 1,700 utt. & ✗ \\
AVSD-Arabic \cite{elhaj2021avsdar}       & Video + Audio        & Lip-reading          & 1,100 vids & ✗ \\
\rowcolor{gray!15}
\textbf{DiDeMo-AR (ours)}                & Video + Text         & \textbf{Video Retrieval} & 40,144 caps & \textbf{✓} \\
\bottomrule
\end{tabular}
\end{table*}


\section{Introduction}

The exponential growth of online video has created an urgent demand for accurate retrieval systems that can find relevant moments within long streams of visual content. On YouTube alone, more than 500 hours of video are uploaded every minute \cite{socialshepherd2025}. 

\begin{figure}[t]
    \centering
    \includegraphics[width=\linewidth]{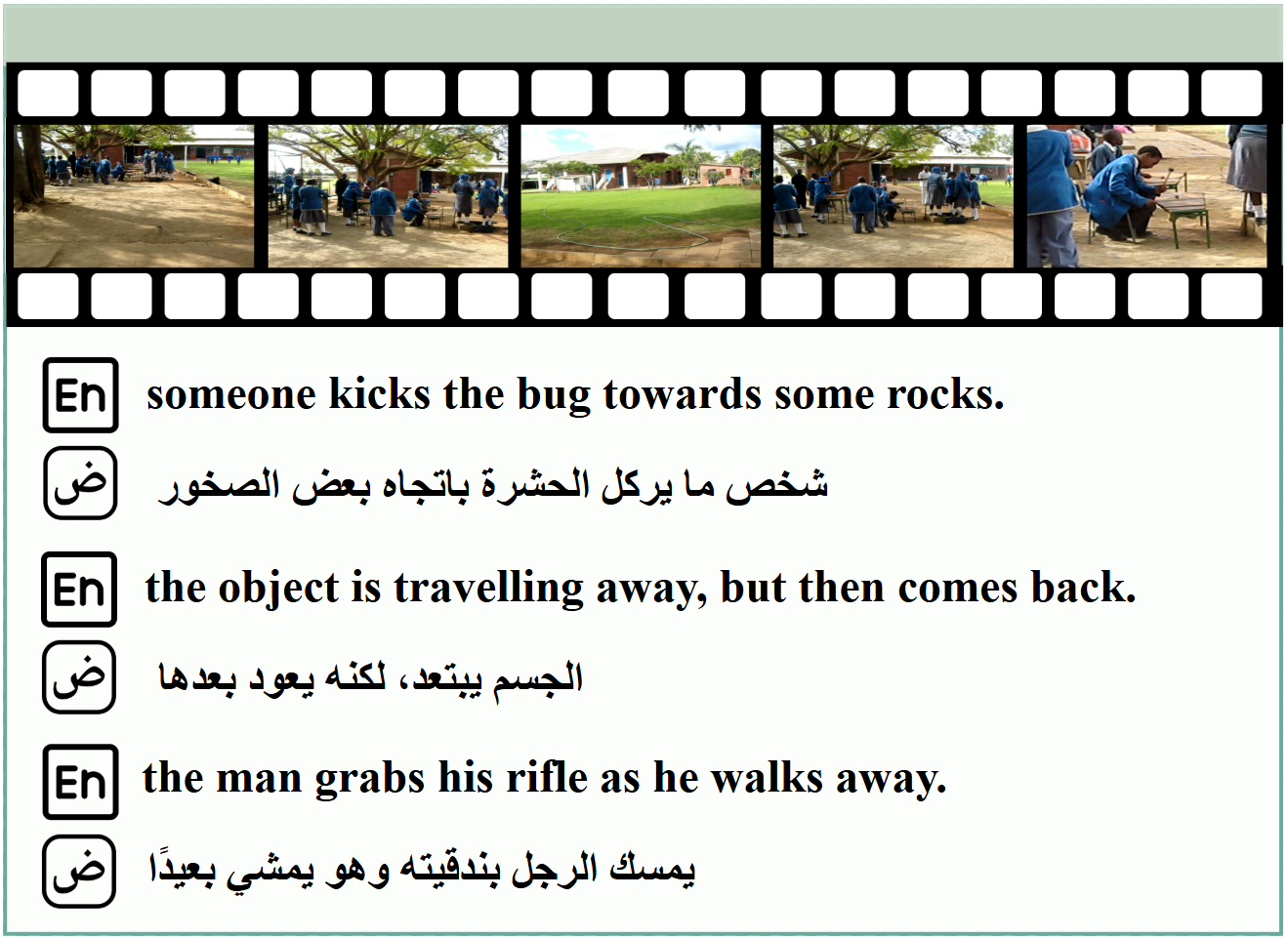}
  \caption{A sample of English captions and their MSA translations for three moments in the same video.}
  \label{fig:frontpage_figs}
\end{figure}

Over the past decade, the research community has released a flood of English-centric benchmarks like DiDeMo \cite{anne2017localizing}, MSR-VTT \cite{xu2016msr}, the bilingual VATEX \cite{wang2019vatex} and the multilingual RUDDER \cite{dabral2021rudder}.

Although these benchmarks have become standard for text-to-video and video-to-text retrieval, all of them completely omit Arabic. Subsequently, Arab researchers are forced to evaluate their retrieval models on English data, literally translated data, or private translations. This slows progress in Arabic multimodal research and questions the reproducibility of their results.

Our work helps fill this gap with a three-stage Large Language Models (LLMs) framework that localizes any non-Arabic retrieval benchmark into Modern Standard Arabic (MSA) with minimal human effort. The framework (i) uses a large language model to translate captions into Modern Standard Arabic, (ii) utilizes a second LLM to automatically flag lexical, grammatical, and formatting errors, and (iii) sends only flagged samples to expert annotators for final verification. The workflow has been applied to the Distinct Describable Moments corpus (DiDeMo), resulting in \textbf{DiDeMo-AR}, the first Arabic video retrieval benchmark, consisting of 10,464 videos and 40,144 fluent Arabic descriptions. We further contribute the first systematic taxonomy of LLM translation errors for Arabic benchmark creation, intended as a reusable checklist for future translation efforts.

To ensure that localization preserves the original benchmark's difficulty, we finetune a \textit{Contrastive Language-Image Pre-training} (CLIP) baseline \cite{radford2021learning} that uses a \textit{Vision Transformer} (ViT-B/16 and ViT-B/32) image encoder \cite{dosovitskiy2020vit} and a \textit{Masked and Permuted Pre-training} (MPNet) text encoder \cite{song2020mpnet}, optimized with the symmetric InfoNCE contrastive loss \cite{oord2018infonce}, on both the English and Arabic variants of DiDeMo. Although Arabic has a complex word structure, the model shows only a $\approx$ 3-point drop in Recall@1 (R@1, higher is better). This result suggests that LLM-based translation, combined with light expert correction, can preserve benchmark difficulty without requiring language-specific pre-training.

We believe this workflow, benchmark, and error analysis will help guide future Arabic benchmark localization research.


\section{Background \& Related Work}

Early attempts to translate multimodal datasets relied either on direct machine translation of English captions or on small teams of human annotators. The MSVD-Indonesian corpus \cite{hendria2023msvd}, for example, was created by translating the original MSVD sentences into Indonesian with Google Translate and then finetuning a CLIP baseline. VATEX offers English-Chinese captions produced by human experts, but no Arabic version, and its captions are sentence-level rather than moment-level \cite{wang2019vatex}. RUDDER combines Google-translated captions with expert annotations and adds five additional languages, yet still omits Arabic entirely \cite{dabral2021rudder}. None of these projects publishes a detailed taxonomy of translation errors, so their contributions remain dataset-specific and provide little guidance for researchers who intend to localize new benchmarks.

Table~\ref{tab:retrieval_benchmarks} lists the retrieval benchmarks that have driven progress during the last decade. Corpora such as MSR-VTT \cite{xu2016msr} and LSMDC \cite{rohrbach2015dataset} are clip-based and English only. DiDeMo \cite{anne2017localizing} introduced moment-level ground-truth in $\sim$10k unedited Flickr videos, followed by ActivityNet Captions, which applies the same idea to long YouTube clips \cite{caba2015activitynet}.

Table \ref{tab:retrieval_benchmarks} highlights a simple fact: not one public retrieval benchmark offers Modern Standard Arabic (MSA) captions, and only two (DiDeMo, ActivityNet) provide moment-level ground truth.

Looking into Arabic multimodal benchmarks, it can be seen that such benchmarks exist but they target tasks very different from retrieval. MGB-3 focuses on broadcast speech and automatic speech recognition \cite{ali2017speech}. MASC provides more than 1,000 hours of YouTube audio for large-scale ASR experiments, again without video captions \cite{al2023masc}. Amd’SaEr utilizes short YouTube clips for sentiment and emotion recognition \cite{haouhat2023towards} . Large text corpora such as ArabicaQA push reading comprehension research forward \cite{abdallah2024arabicaqa} , yet contain no video. Table~\ref{tab:arabic_multimodal} summarizes the information from these datasets.
To the best of our knowledge, \textbf{DiDeMo-AR} is therefore
the first publicly released benchmark that pairs Arabic
sentences with temporally grounded video moments.

\begin{figure}[t]  
    \centering
    \includegraphics[width=\linewidth]{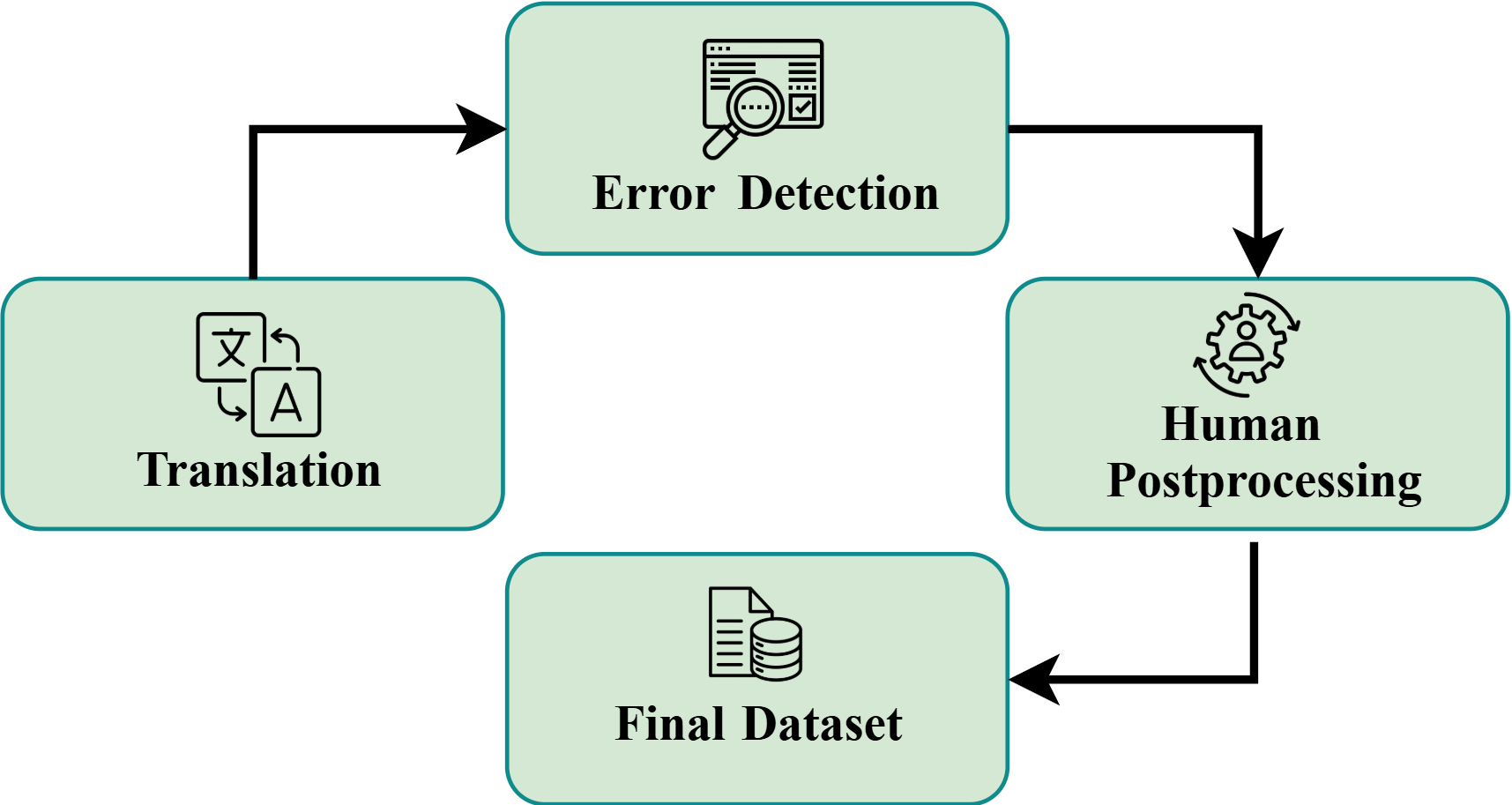}
    \caption{\textsc{AutoArabic} three-stage localization workflow: translation, error detection, and human post-editing.}
    \label{fig:pipeline_diagram}
\end{figure}

\begin{table*}[t]
\centering
\footnotesize
\renewcommand{\arraystretch}{1.2}
\caption{Error categories identified by the automated detector and addressed through manual post-editing.}
\begin{tabular}{@{}lp{5.2cm}p{7.3cm}@{}}
\toprule
\textbf{Error Type} & \textbf{Definition} & \textbf{Example (English / Arabic)} \\
\midrule
\textit{Lexical} &
Selection of uncommon or overly formal words instead of familiar alternatives. &
EN: \textit{first time we see an otter swim by} \\
& & AR-poor: \textarabic{هذه أول مرة نرى فيها \textbf{قضاعة} تسبح.} \\
& & AR-improved: \textarabic{هذه أول مرة نرى \textbf{ثعلب الماء} يسبح.} \\
\addlinespace

\textit{Literal} &
Word-for-word structural translation that produces unnatural Arabic phrasing. &
EN: \textit{The man raises onto his knees to crawl.} \\
& & AR-poor: \textarabic{يرفع الرجل جذعه ليستند على ركبته زحفاً.} \\
& & AR-improved: \textarabic{ينهض الرجل على ركبتيه ليزحف.} \\
\addlinespace

\textit{Hallucination} &
Addition of content not present in the original English text. &
EN: \textit{The girl starts speaking.} \\
& & AR-poor: \textarabic{الفتاة تبدأ بالتحدث \textbf{باللغة العربية}.} \\
& & AR-improved: \textarabic{الفتاة تبدأ بالتحدث.} \\
\addlinespace

\textit{Tense Shift} &
Incorrect temporal rendering of present actions in past tense. &
EN: \textit{Person in black exits frame to left.} \\
& & AR-poor: \textarabic{\textbf{خرج} الشخص ذو اللباس الأسود من المشهد نحو اليسار.} \\
& & AR-improved: \textarabic{\textbf{يخرج} الشخص ذو اللباس الأسود من المشهد نحو اليسار.} \\
\addlinespace

\textit{Loanword} &
Inconsistent use of transliterated terms versus established Arabic equivalents. &
EN: \textit{The camera zooms up on the players.} \\
& & AR-poor: \textarabic{تقترب \textbf{الكاميرا} بالتكبير على اللاعبين.} \\
& & AR-improved: \textarabic{تقترب \textbf{آلة التصوير} بالتكبير على اللاعبين.} \\
\addlinespace

\textit{Diacritics} &
Inconsistent application of diacritical marks across words and captions. &
EN: \textit{The gentleman puts his left arm under his right arm.} \\
& & AR-poor: \textarabic{يضعُ الرَّجُلُ ذِرَاعَهُ الْيُسْرَى تحت ذِرَاعِهِ الْيُمْنَى.} \\
& & AR-improved: \textarabic{يضع الرجل ذراعه اليسرى تحت ذراعه اليمنى.} \\
\bottomrule
\end{tabular}

\label{tab:error_definitions}
\end{table*}


\section{The \textsc{AutoArabic} Framework}

Figure 2 shows \textsc{AutoArabic}, a three-stage framework that can turn any English video-text benchmark into Modern Standard Arabic (MSA). 
In this section we describe the framework in general terms. Its output for DiDeMo is analyzed in the next sections.

First, every English caption is sent to Gemini 2.0 Flash  \cite{google2025geminiflash} with this prompt:

\begin{verbatim}
"You will receive an English sentence that 
serves as a caption for a short video clip. 
Your task is to translate this caption 
into Modern Standard Arabic while ensuring 
that the translation remains suitable  
and appropriate as a caption.
The English caption: {caption}
Arabic caption:"
\end{verbatim}

Gemini is run with \texttt{temperature=0.7} and \texttt{top-p=1.0}.  
Next, each Arabic output is processed by GPT-4o \cite{openai2025gpt4o} for automatic error detection, tagging six categories: \texttt{lexical}, \texttt{literal}, \texttt{hallucination}, \texttt{tense\_shift}, \texttt{loanword}, and \texttt{diacritics} (summarized in Table~\ref{tab:error_definitions}).    

Finally, captions flagged by the detector are reviewed by five native-speaker annotators. Although the framework supports selective post-editing, we performed a full revision in this study, where annotators reviewed every caption rather than only the flagged ones. 
We compared the error detection performance of the LLM against that of the annotators and found that the LLM successfully identified over 97\% of the actual mistakes.  

Using these reviewed captions, we evaluated caption quality under \emph{three post-editing budgets}: (i) Raw LLM output (zero), (ii) Fix only LLM-flagged (few), and (iii) Fix all (full). Results show that performance improves monotonically with greater post-editing (zero $\!\rightarrow$ few $\!\rightarrow$ full), while the raw LLM output remains usable.  

It is worth mentioning that the framework is provider-agnostic: the prompting, validation, and post-processing steps do not depend on a specific API and can be run with open or proprietary LLMs. In this paper, we used high-performing commercial models to maximize one-time localization quality. 

Additionally, diacritics themselves are \emph{not} errors; \emph{inconsistency} across samples is. We intentionally did not constrain diacritics in the translation prompt to observe natural model behavior, then enforced uniformity post hoc via deterministic stripping. 


\begin{figure*}[!ht]
  \centering
  \colorbox{gray!10}{%
  \begin{minipage}{0.95\textwidth}
    \centering
    \vspace{0.5ex}
    \textbf{Word-Count Distributions}
    \vspace{0.5ex}
    
    \begin{subfigure}[t]{0.48\textwidth}
      \centering
      \includegraphics[width=\linewidth]{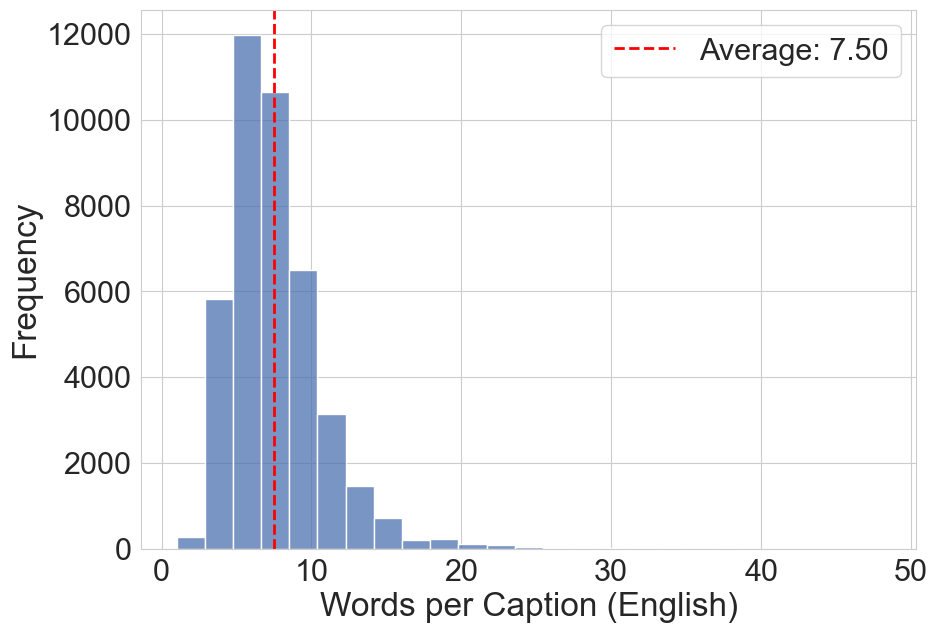}
      \caption{English word‑count distribution}
    \end{subfigure}\hfill
    \begin{subfigure}[t]{0.48\textwidth}
      \centering
      \includegraphics[width=\linewidth]{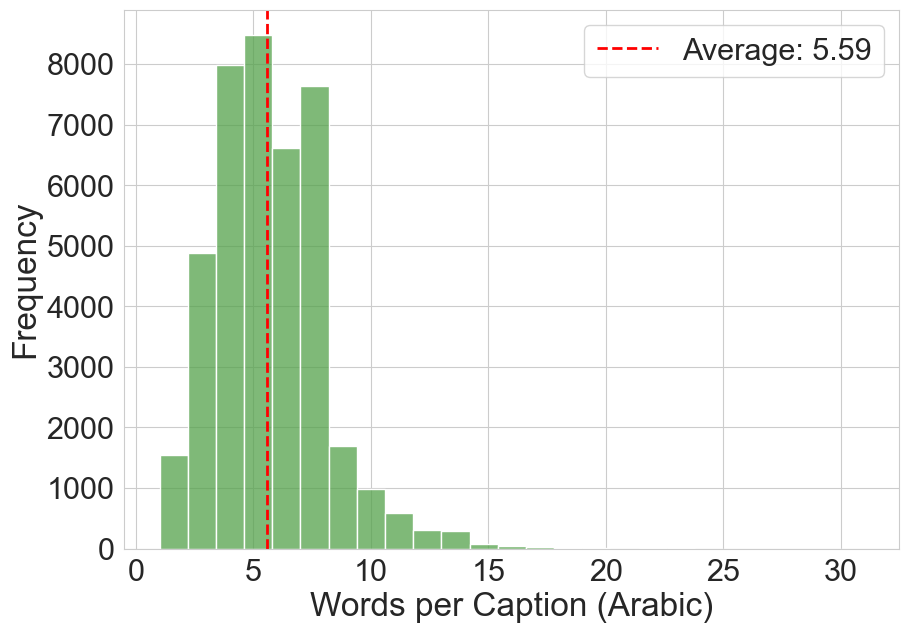}
      \caption{Arabic word‑count distribution}
    \end{subfigure}
    \vspace{0.5ex}
  \end{minipage}%
  }
  \vspace{2ex}
  \colorbox{blue!5}{%
  \begin{minipage}{0.95\textwidth}
    \centering
    \vspace{0.5ex}
    \textbf{Part-of-Speech Distributions}
    \vspace{0.5ex}
    
    \begin{subfigure}[t]{0.33\textwidth}
      \centering
      \includegraphics[width=\linewidth]{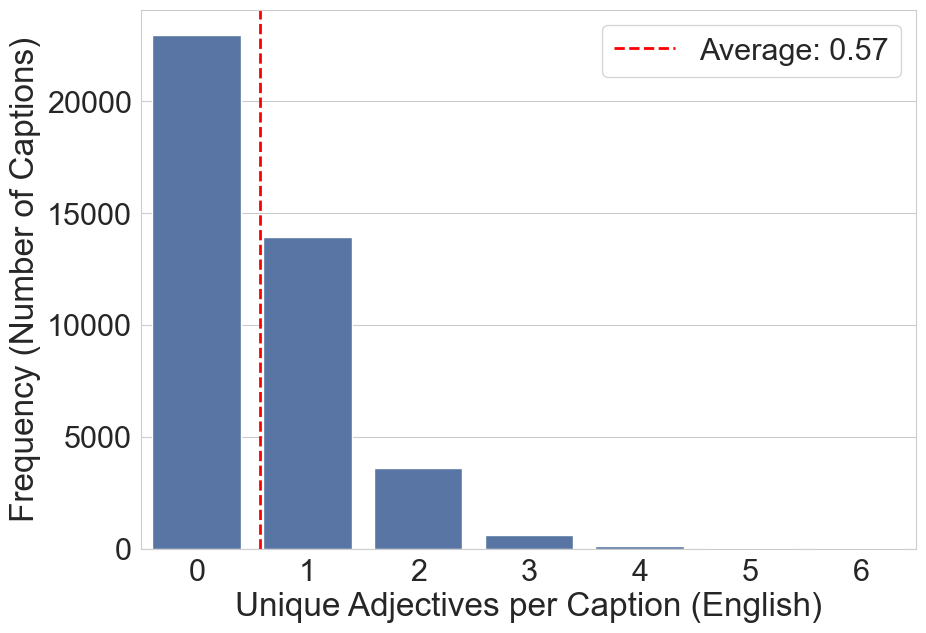}
      \caption{EN adjectives}
    \end{subfigure}\hfill
    \begin{subfigure}[t]{0.33\textwidth}
      \centering
      \includegraphics[width=\linewidth]{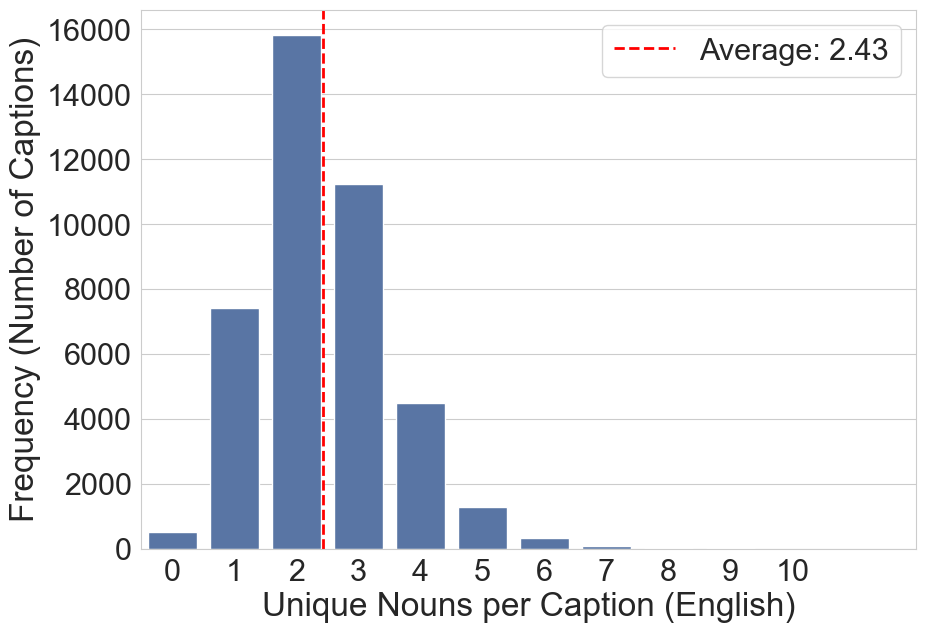}
      \caption{EN nouns}
    \end{subfigure}\hfill
    \begin{subfigure}[t]{0.33\textwidth}
      \centering
      \includegraphics[width=\linewidth]{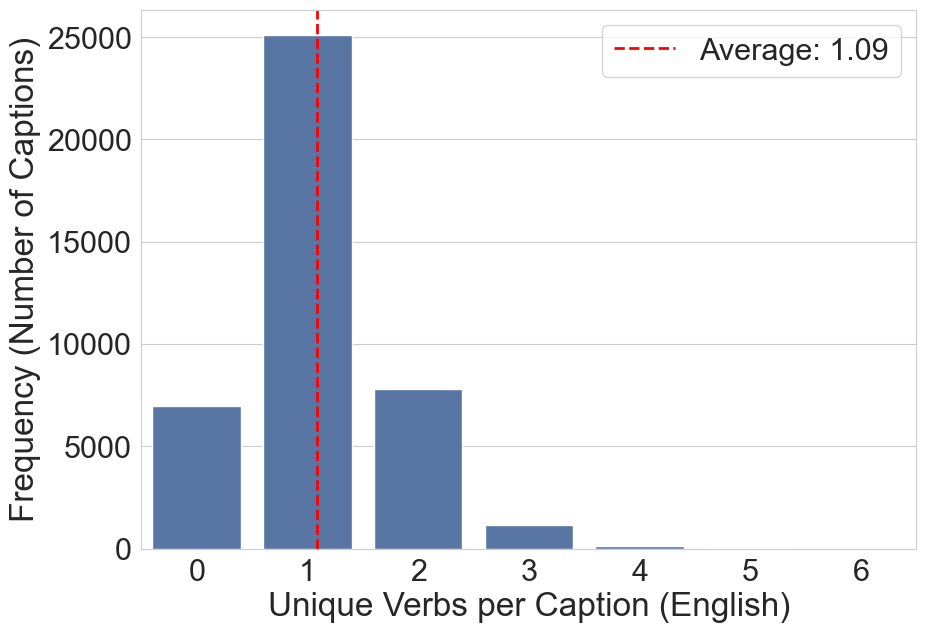}
      \caption{EN verbs}
    \end{subfigure}
    \vspace{1.5ex}
    \begin{subfigure}[t]{0.33\textwidth}
      \centering
      \includegraphics[width=\linewidth]{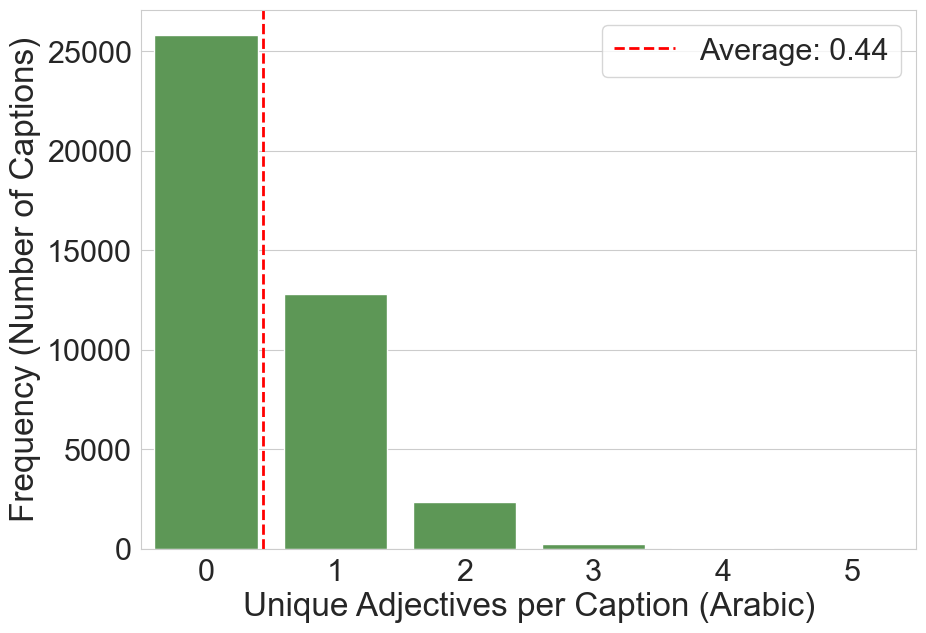}
      \caption{AR adjectives}
    \end{subfigure}\hfill
    \begin{subfigure}[t]{0.33\textwidth}
      \centering
      \includegraphics[width=\linewidth]{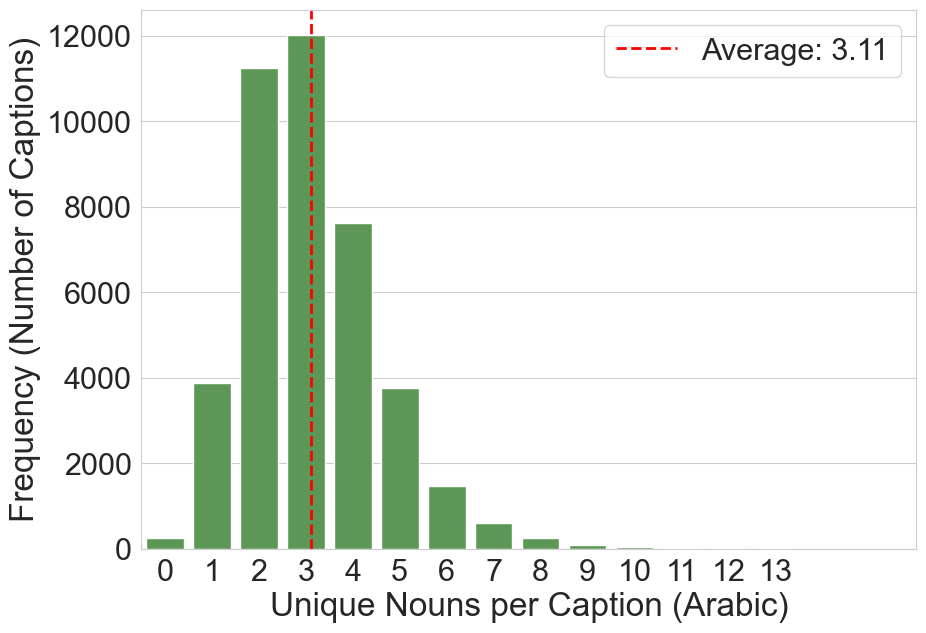}
      \caption{AR nouns}
    \end{subfigure}\hfill
    \begin{subfigure}[t]{0.33\textwidth}
      \centering
      \includegraphics[width=\linewidth]{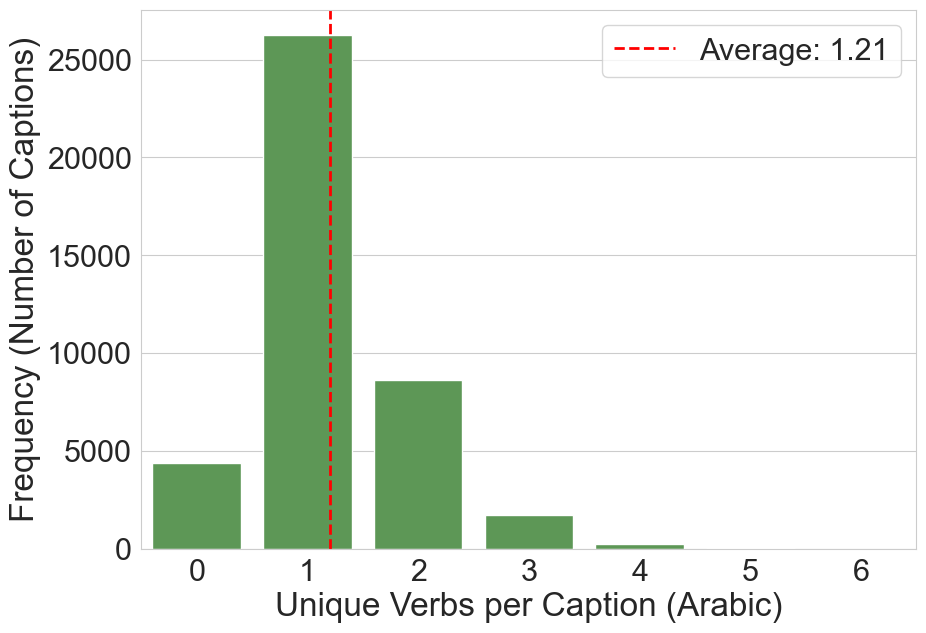}
      \caption{AR verbs}
    \end{subfigure}
    \vspace{0.5ex}
  \end{minipage}%
  }
  \caption{
    \textbf{Top}: Word‑count distributions per caption for English (left) and Arabic (right) in DiDeMo vs. DiDeMo-AR.
    \textbf{Middle}: Distributions of unique adjectives, nouns, and verbs per caption in English (DiDeMo).
    \textbf{Bottom}: Same distributions for Arabic (DiDeMo-AR).
  }
  \label{fig:all_distributions}
\end{figure*}

\begin{figure*}[t]  
    \centering
    \begin{subfigure}[t]{0.49\textwidth}
        \includegraphics[width=\linewidth]{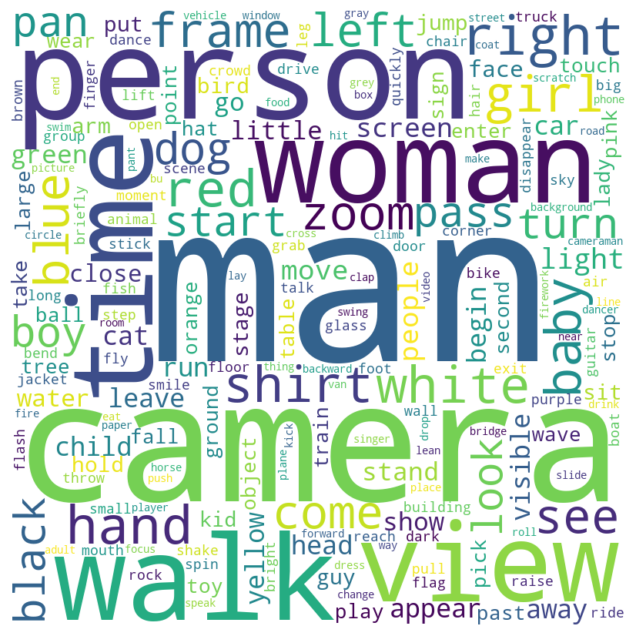}
        \caption{English}
    \end{subfigure}
    \hfill
    \begin{subfigure}[t]{0.49\textwidth}
        \includegraphics[width=\linewidth]{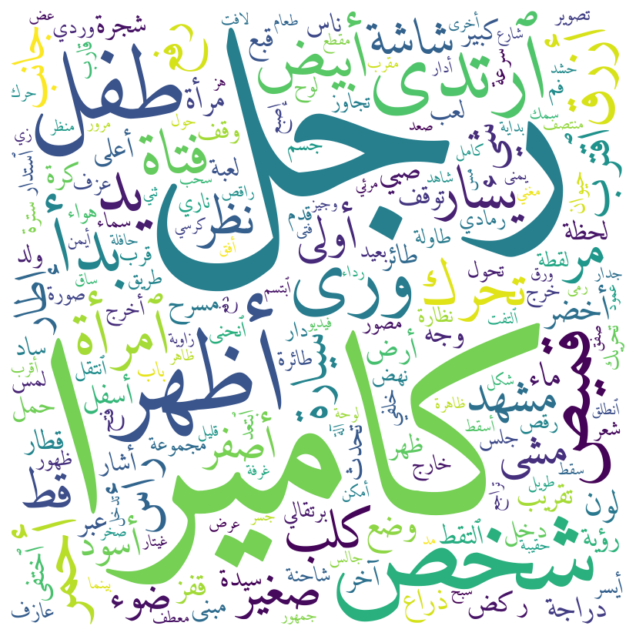}
        \caption{Arabic}
    \end{subfigure}
    \caption{Word cloud visualization in English and Arabic captions.}
    \label{fig:word_clouds}
\end{figure*}

\section{The \textsc{DiDeMo-AR} Dataset}\label{sec:didero_ar}

The Distinct Describable Moments (DiDeMo) dataset \cite{anne2017localizing} is one of the largest and most diverse datasets for the temporal localization of events in videos given natural language descriptions. The videos are collected from Flickr and each video is trimmed to a maximum of 30 seconds. The videos in the dataset are divided into 5-second segments to reduce the complexity of annotation. The dataset is split into training, validation and test sets containing 8,395, 1,065 and 1,004 videos respectively. The dataset contains a total of 26,892 moments and one moment could be associated with descriptions from multiple annotators. The total number of captions in DiDeMo is 40,144. The descriptions in DiDeMo dataset are detailed and contain camera movement, temporal transition indicators, and activities. Moreover, the descriptions in DiDeMo are verified so that each description refers to a single moment.

Applying the translation framework to DiDeMo yields \textbf{DiDeMo-AR} with the same 10,464 videos and 26,892 moments, but now 40,144 fluent MSA captions.
Arabic captions are slightly shorter, 5.6 words on average versus 7.5 in English.
Figure \ref{fig:all_distributions} plots the word-per-caption distribution for both languages on the top, while  Figure \ref{fig:word_clouds} visualizes the most frequent content words. It can be seen that the most common words in English also appear in the Arabic figure with nearly the same size, indicating consistent translation and semantic mapping across languages. 

Table~\ref{tab:lex_pos} reports unique $n$-gram and POS counts.
While Arabic and English share a similar 1-gram vocabulary count, the counts diverge as we move to longer $n$-gram. Regarding POS tokens, Arabic shows a smaller set of distinct POS tokens compared to English. Achieving performance close to the English baseline with a smaller lexical set shows the concise expressive power of Arabic.

During manual revision, we logged the errors found in every caption. Their distribution is shown in Table \ref{tab:error_breakdown}, where error rate denotes the percentage of captions containing $\geq\!1$ instance of the category (totals can exceed 100\% because a caption may contain multiple categories).
The most frequent issue is inconsistent use of diacritics (some captions contain full diacritics while others have none) accounting for 27.8\% of the entire dataset.
Loanword handling ranks second (12.7\%), followed by tense shifts (3.4\%).
Literal translations, rare lexical choices, and hallucinations together occur in fewer than 5\% of captions.

\captionsetup[table]{skip=5pt}
\begin{table}[H]
\centering
\footnotesize
\caption{Unique $n$-grams and POS-tag counts in DiDeMo vs.\ DiDeMo-AR.}
\begingroup
\renewcommand{\arraystretch}{2.15}        
\setlength{\aboverulesep}{0.8ex}          
\setlength{\belowrulesep}{0.8ex}          
\begin{tabular}{@{}lrrrr@{}}
\toprule
\textbf{Language} & 1-gram & 2-gram & 3-gram & 4-gram \\
\midrule
English & 5,358 & 67,698 & 140,387 & 163,841 \\
\addlinespace[0.4ex]
Arabic  & 5,205 & 75,904 & 151,943 & 176,369 \\
\midrule
\textbf{POS} & verbs & nouns & adj. & adv. \\
\midrule
English & 1,320 & 3,605 & 891 & 333 \\
\addlinespace[0.4ex]
Arabic  & 1,145 & 2,822 & 713 & 17 \\
\bottomrule
\end{tabular}
\endgroup
\label{tab:lex_pos}
\end{table}

\captionsetup[table]{skip=5pt}
\begin{table}[H]
\centering
\footnotesize
\setlength{\tabcolsep}{15pt}
\caption{\textbf{Top}: exclusive single-error rates on the DiDeMo-AR dataset. \textbf{Bottom}: distribution of captions that shows multiple error types simultaneously.}
\begin{tabular}{@{}p{6cm}r@{}}
    \toprule
    \textbf{Error Type} & \textbf{\%} \\
    \midrule
    Diacritics                 & 27.8 \\
    Loanwords                  & 12.7 \\
    Literal / weak phrasing    &  5.0 \\
    Tense shift                &  3.4 \\
    Hallucination              &  1.8 \\
    \midrule
    Total error rate (overlapped) & \textbf{41.7} \\
    \midrule[\heavyrulewidth]
    \textbf{Overlap Type} & \textbf{\%} \\
    \midrule
    Loanword + Diacritics           & 7.1 \\
    Tense shift + Diacritics        & 1.6 \\
    Tense shift + Loanword          & 0.4 \\
    Tense + Loan + Diac.            & 0.1 \\
    \bottomrule
  \end{tabular}

\label{tab:error_breakdown}
\vspace{-1.2ex}
\end{table}

Combinations of these errors occur in a small portion of the data, with the most common overlap being loanword + diacritics (7.1\%), followed by tense shift + diacritics (1.6\%) and tense shift + loanword (0.4\%). Only 0.1\% of captions show more than two error types simultaneously.

Annotators resolved the diacritics issue by stripping all diacritics, ensuring consistent style across the corpus. For loanwords, annotators kept terms that are widely used in Modern Standard Arabic, for example, \textarabic{"كاميرا"} is already commonly used and preferred over the more formal \textarabic{"آلة التصوير"} 
All remaining errors were manually corrected.

We also noticed that Gemini occasionally inserts the phrase "\textarabic{باللغة العربية}" ("in Arabic") at the end of a few captions.
This seems to happen when the model treats the final words of the prompt as part of the source text.
Annotators removed these additions manually, but future work should craft prompts carefully, by ensuring source text and prompt are clearly distinguishable, to avoid similar issues.

Finally, Gemini sometimes translates only part of a caption if it contains verbs such as "is shown" or "appears."
For example:

\begin{itemize}\setlength\itemsep{2pt}
\item \textbf{English:} "The words 'the gossip' are shown first."
\item \textbf{Incorrect AR:} \textarabic{النميمة}
\item \textbf{Correct AR:} \textarabic{تظهر كلمة "النميمة" أولاً.}
\end{itemize}

These partial translations were also fixed during post-editing.

We also experimented with different temperatures values to test the translation’s sensitivity to the decoding settings. Temperature primarily controls sampling randomness, where higher values encourage more lexical variety, while lower values make outputs more deterministic.
We tested $\{0.0, 0.1, \dots, 1.0\}$, but the outputs differed only in minor synonym choices (e.g., \textarabic{تلوّح} vs.\ \textarabic{تلوح}), confirming that Gemini's Arabic translation remains stable across all settings.

Some noise also stems from the English side of DiDeMo itself. A few captions are simply ambiguous, for instance \ "they zoom back in at the end" gives no clue who performs the action, so even a perfect translator cannot disambiguate it. On the other hand, most plain grammar or spelling mistakes in the source are corrected automatically: "a car drive under and overpass" is translated fluently as "\textarabic{تمرُّ سيارةٌ تحت جسرٍ علوي}". Gemini, likewise, resolves DiDeMo shortcuts such as "ppl", which was translated to "\textarabic{الناس}". In short, some inherited flaws remain, but many are silently repaired in the Arabic version, and although there is some translation noise, Gemini's raw output is already usable. Diacritics can be removed programmatically, and other post-editing fixes are needed for only \textbf{22.9\%} of captions (after diacritic stripping).

\begin{figure*}[!ht]  
    \centering
    \begin{subfigure}[b]{0.49\linewidth}
        \includegraphics[width=\linewidth]{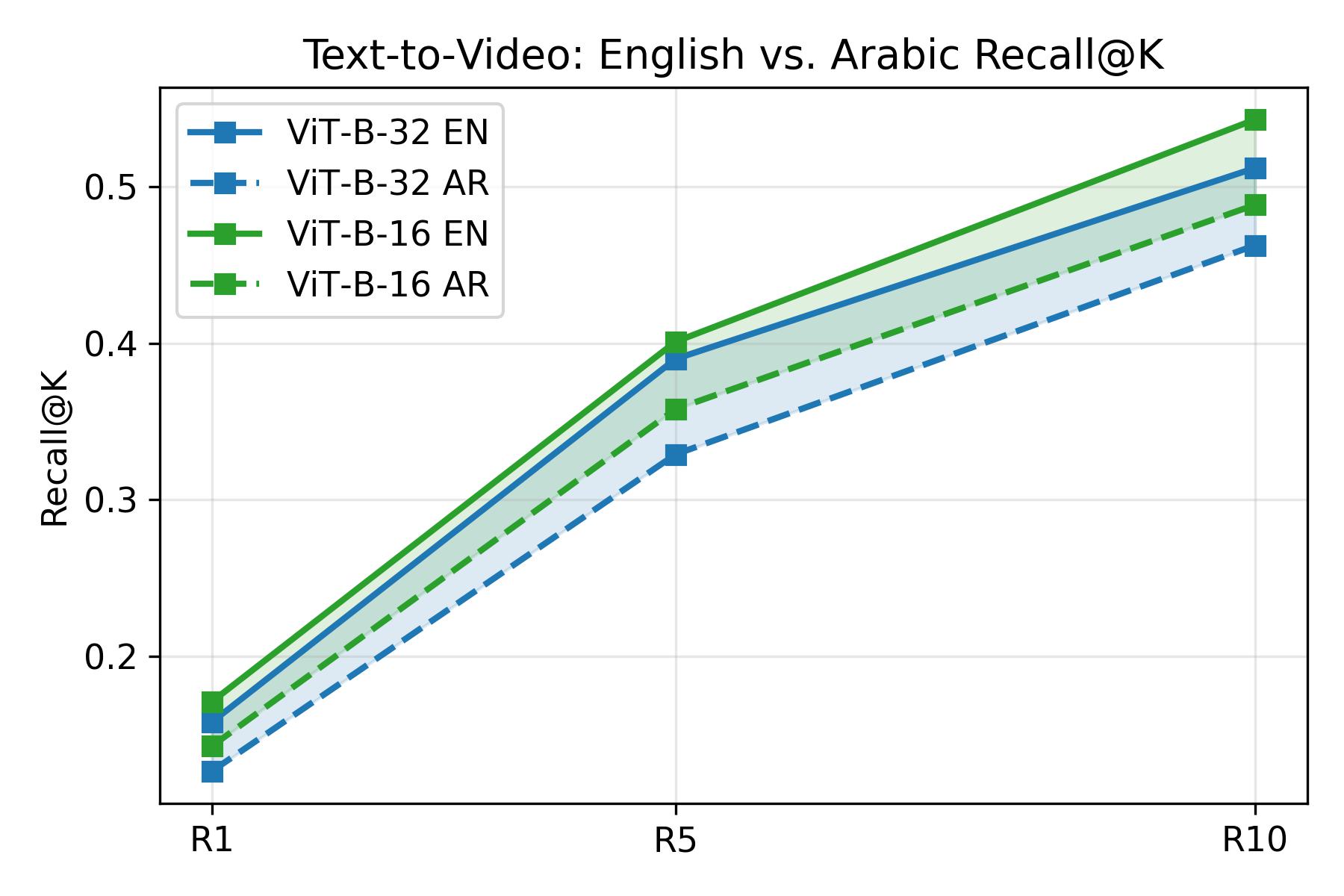}
        \caption{Text-to-Video Retrieval}
        \label{fig:ttv}
    \end{subfigure}
    \hfill
    \begin{subfigure}[b]{0.49\linewidth}
        \includegraphics[width=\linewidth]{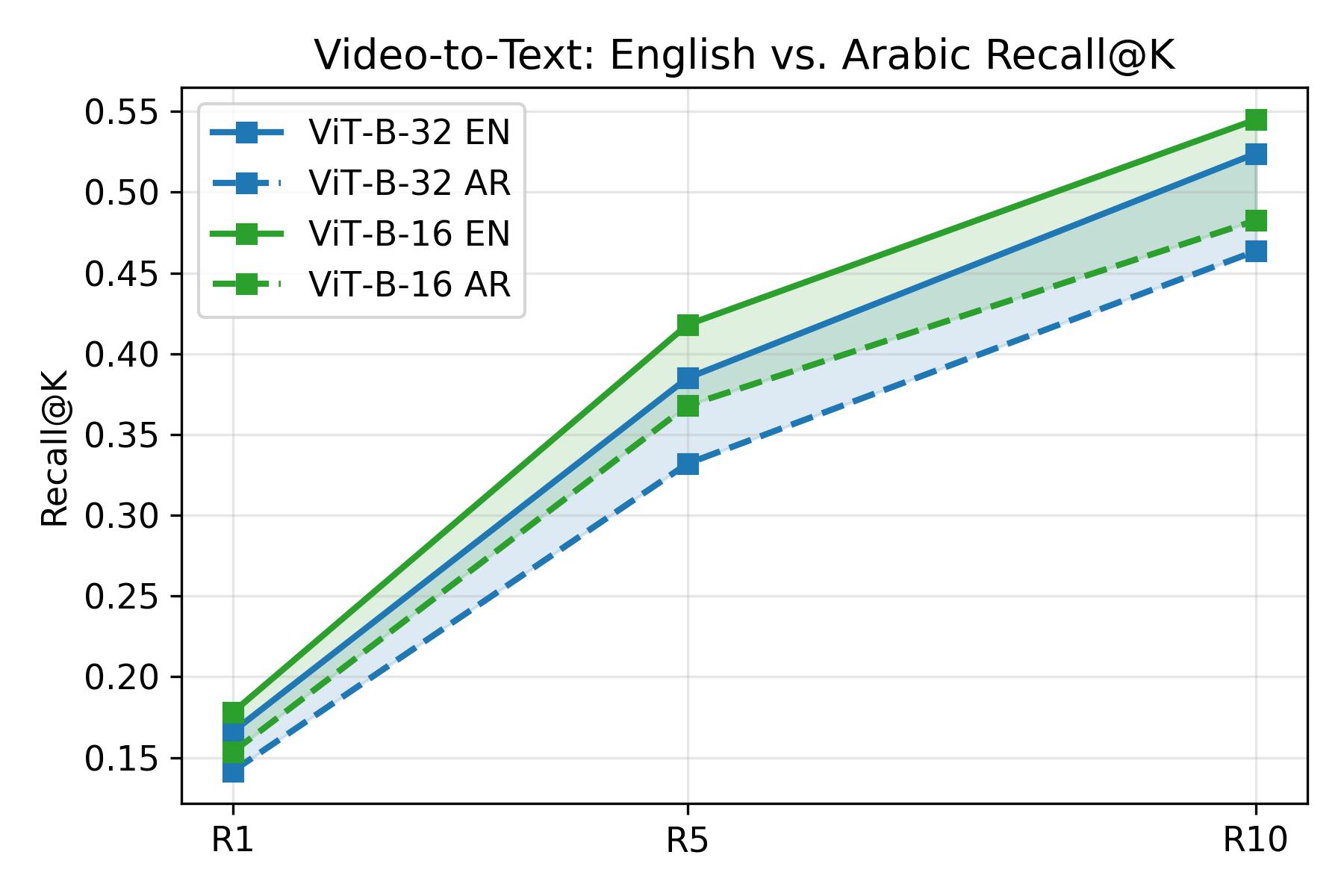}
        \caption{Video-to-Text Retrieval}
        \label{fig:vtt}
    \end{subfigure}
    \caption{English vs. Arabic performance comparison in text-video and video-text retrieval (Recall@K).}
    \label{fig:ttv_vtt}
\end{figure*}

\captionsetup[table*]{skip=9pt}
\begin{table*}[!t]
  \centering
  \small
  \setlength{\tabcolsep}{10pt}
  \renewcommand{\arraystretch}{1.2}
  \caption{Text-to-Video retrieval performance on DiDeMo test split. \textcolor{red}{$\Delta$}: the Arabic-English performance gap.}
  \begin{tabular}{@{}llccccc@{}}
    \toprule
    \textbf{Model} & \textbf{Lang.} & \textbf{R@1} $\uparrow$ & \textbf{R@5} $\uparrow$ & \textbf{R@10} $\uparrow$ & \textbf{MedR} $\downarrow$ & \textbf{MeanR} $\downarrow$\\
    \midrule
    \multirow{2}{*}{\textbf{ViT-B-32 + MPNet}} 
    & EN & \textbf{0.158} & \textbf{0.390} & \textbf{0.512} & \textbf{10} & \textbf{48.2}\\
    & AR & 0.127 \textcolor{red}{\scriptsize($\Delta$-0.031)} & 
           0.329 \textcolor{red}{\scriptsize($\Delta$-0.061)} & 
           0.463 \textcolor{red}{\scriptsize($\Delta$-0.049)} & 
           13 \textcolor{red}{\scriptsize($\Delta$+3)} & 
           55.7 \textcolor{red}{\scriptsize($\Delta$+7.5)}\\
    \addlinespace
    \multirow{2}{*}{\textbf{ViT-B-16 + MPNet}}
    & EN & \textbf{0.171} & \textbf{0.401} & \textbf{0.543} & \textbf{8} & \textbf{45.9}\\
    & AR & 0.143 \textcolor{red}{\scriptsize($\Delta$-0.028)} & 
           0.358 \textcolor{red}{\scriptsize($\Delta$-0.043)} & 
           0.489 \textcolor{red}{\scriptsize($\Delta$-0.055)} & 
           11 \textcolor{red}{\scriptsize($\Delta$+3)} & 
           50.6 \textcolor{red}{\scriptsize($\Delta$+4.7)}\\
    \bottomrule
  \end{tabular}
  
  \label{tab:ttv_results}
\end{table*}

\captionsetup[table*]{skip=9pt}
\begin{table*}[!t]
  \centering
  \small
  \setlength{\tabcolsep}{10pt}
  \renewcommand{\arraystretch}{1.2}
  \caption{Video-to-Text retrieval performance on DiDeMo test split. \textcolor{red}{$\Delta$}: the Arabic-English performance gap.}
  \begin{tabular}{@{}llccccc@{}}
    \toprule
    \textbf{Model} & \textbf{Lang.} & \textbf{R@1} $\uparrow$ & \textbf{R@5} $\uparrow$ & \textbf{R@10} $\uparrow$ & \textbf{MedR} $\downarrow$ & \textbf{MeanR} $\downarrow$\\
    \midrule
    \multirow{2}{*}{\textbf{ViT-B-32 + MPNet}} 
    & EN & \textbf{0.166} & \textbf{0.385} & \textbf{0.524} & \textbf{9} & \textbf{48.3}\\
    & AR & 0.142 \textcolor{red}{\scriptsize($\Delta$-0.024)} & 
           0.332 \textcolor{red}{\scriptsize($\Delta$-0.053)} & 
           0.464 \textcolor{red}{\scriptsize($\Delta$-0.060)} & 
           13 \textcolor{red}{\scriptsize($\Delta$+4)} & 
           54.3 \textcolor{red}{\scriptsize($\Delta$+6.0)}\\
    \addlinespace
    \multirow{2}{*}{\textbf{ViT-B-16 + MPNet}}
    & EN & \textbf{0.178} & \textbf{0.418} & \textbf{0.545} & \textbf{8} & \textbf{44.9}\\
    & AR & 0.154 \textcolor{red}{\scriptsize($\Delta$-0.025)} & 
           0.368 \textcolor{red}{\scriptsize($\Delta$-0.050)} & 
           0.483 \textcolor{red}{\scriptsize($\Delta$-0.062)} & 
           11 \textcolor{red}{\scriptsize($\Delta$+3)} & 
           49.8 \textcolor{red}{\scriptsize($\Delta$+4.9)}\\
    \bottomrule
  \end{tabular}
  
  \label{tab:vtt_results}
\end{table*}

\section{Experiments \& Results}

\subsection{Setup \& Baselines}
\label{sec:setup}

We fine-tune two CLIP backbones \textbf{ViT-B/32} and \textbf{ViT-B/16}, while
\emph{freezing} the vision tower and updating only a 256-d projection head.
The text branch is \texttt{paraphrase-multilingual-mpnet-base-v2}
(768\,d; 110 M parameters).
Training follows a symmetric InfoNCE loss, batch size 64, AdamW
($\mathrm{lr}\!=\!1\mathrm{e}{-4}$, weight-decay $1\mathrm{e}{-2}$) and
runs for six epochs on one A100-80 GB.
Input videos are down-sampled to eight uniformly spaced frames
($224\times224$).  
We train identical scripts on the original English captions and on the new
Arabic set, so any gap is purely linguistic.
Our CLIP baseline is deliberately lightweight. Its role is to verify that the Arabic variant remains comparably difficult, not to exhaustively benchmark Arabic video-retrieval models.

\subsection{Overall Retrieval Scores}
\label{sec:overall}

Tables~\ref{tab:ttv_results} and~\ref{tab:vtt_results}
report Recall@K, Median Rank, and Mean Rank on the DiDeMo test split.
Despite Arabic captions being 25\% shorter, the absolute drop is small:
\mbox{$\Delta$R@1 $\!<$ 3 pp} for both ViT backbones in
\emph{text-to-video} and \emph{video-to-text} directions.
Median rank increases by three to four positions on average, but still stays
below 15.

Figure \ref{fig:ttv_vtt} overlays English and Arabic curves.
The shaded area highlights the gap. It never exceeds 0.07 at R@10.
This shows that performance gaps remain nearly parallel across R@1, 5, 10.

Using the fully post-edited Arabic captions, a frozen CLIP backbone recovers 85-90\% of its English Recall@10. 
This confirms that \emph{metric localization} using our framework preserves benchmark difficulty without extra Arabic pre-training, with most of the English retrieval strength transferring directly to Arabic.

\captionsetup[table*]{skip=9pt}
\begin{table*}[!ht]  
  \centering
  \small
  \setlength{\tabcolsep}{10pt}
  \renewcommand{\arraystretch}{1.2}
  \caption{Text-to-Video retrieval across post-editing levels on DiDeMo-AR.}
  \begin{tabular}{@{}llccccc@{}}
    \toprule
    \textbf{Model} & \textbf{Post-Editing} & \textbf{R@1} $\uparrow$ & \textbf{R@5} $\uparrow$ & \textbf{R@10} $\uparrow$ & \textbf{MedR} $\downarrow$ & \textbf{MeanR} $\downarrow$ \\
    \midrule
    \multirow{3}{*}{\textbf{ViT-B-16 + MPNet}}
    & Raw (zero) & 0.1196 & 0.3230 & 0.4676 & 13.0 & 55.9 \\
    & Flagged-only (few) & 0.1316 & 0.3121 & 0.4556 & 12.0 & 55.3 \\
    & Fix all (full) & \textbf{0.1426} & \textbf{0.3579} & \textbf{0.4885} & \textbf{11.0} & \textbf{50.6} \\
    \addlinespace
    \multirow{3}{*}{\textbf{ViT-B-32 + MPNet}}
    & Raw (zero) & 0.1176 & 0.3270 & 0.4636 & 13.0 & 55.2 \\
    & Flagged-only (few) & 0.1157 & 0.3310 & 0.4646 & 13.0 & 54.9 \\
    & Fix all (full) & 0.1266 & 0.3290 & 0.4626 & 13.0 & 55.7 \\
    \bottomrule
  \end{tabular}
  
  \label{tab:ttv_budget}
\end{table*}

\captionsetup[table*]{skip=9pt}
\begin{table*}[!ht] 
  \centering
  \small
  \setlength{\tabcolsep}{10pt}
  \renewcommand{\arraystretch}{1.2}
  \caption{Video-to-Text retrieval across post-editing levels on DiDeMo-AR.}
  \begin{tabular}{@{}llccccc@{}}
    \toprule
    \textbf{Model} & \textbf{Post-Editing} & \textbf{R@1} $\uparrow$ & \textbf{R@5} $\uparrow$ & \textbf{R@10} $\uparrow$ & \textbf{MedR} $\downarrow$ & \textbf{MeanR} $\downarrow$ \\
    \midrule
    \multirow{3}{*}{\textbf{ViT-B-16 + MPNet}}
    & Raw (zero) & 0.1306 & 0.3519 & \textbf{0.4835} & 11.0 & 53.2 \\
    & Flagged-only (few) & 0.1236 & 0.3500 & 0.4726 & 12.0 & 54.2 \\
    & Fix all (full) & \textbf{0.1535} & \textbf{0.3679} & 0.4826 & \textbf{11.0} & \textbf{49.8} \\
    \addlinespace
    \multirow{3}{*}{\textbf{ViT-B-32 + MPNet}}
    & Raw (zero) & 0.1296 & 0.3420 & 0.4646 & 12.0 & 54.6 \\
    & Flagged-only (few) & 0.1286 & 0.3450 & 0.4646 & 13.0 & 54.4 \\
    & Fix all (full) & 0.1416 & 0.3320 & 0.4636 & 13.0 & 54.3 \\
    \bottomrule
  \end{tabular}
  \label{tab:vtt_budget}
\end{table*}

\subsection{Effect of Post-Editing Effort}\label{sec:budget}  
To understand how human post-editing impacts retrieval performance, we evaluate three levels of manual correction on Arabic captions:
\begin{itemize}[noitemsep, topsep=0pt, leftmargin=*]
    \item \textbf{Raw (zero):} Direct LLM output without human intervention.
    \item \textbf{Flagged-only (few):} Corrections applied only to LLM-flagged captions.
    \item \textbf{Fix all (full):} Comprehensive manual review and correction of all captions.
\end{itemize}

Tables~\ref{tab:ttv_budget} and~\ref{tab:vtt_budget} show that even raw LLM translations achieve reasonable performance. However, increasing post-editing effort yields consistent improvements, with full correction typically providing $\approx$ 2 percentage points gains in R@1 across both retrieval directions.

Notably, if raw translations already work, then benchmark replication becomes language-agnostic, no per-language retraining or major human effort required, provided a capable translation LLM.

\subsection{Automated Error-Flagging Quality}\label{sec:judge}  
We evaluate the LLM-based error detector on our human-reviewed dataset. The automated system achieves strong agreement with human annotators: 97\% accuracy and 91\% F1-score (macro-averaged).

Table~\ref{tab:judge} shows the detector performs perfectly on diacritics and achieves high precision for hallucination detection. Tense shifting proves most challenging (F1=0.80), reflecting the complexity of Arabic temporal expressions.

\begin{table}[!ht]  
  \centering
  \small
  \setlength{\tabcolsep}{10pt}
  \renewcommand{\arraystretch}{1.2}
  \caption{Per-class precision, recall, and F1-score of the automated error-flagging system.}
  \begin{tabular}{@{}lccc@{}}
    \toprule
    \textbf{Class} & \textbf{Precision} & \textbf{Recall} & \textbf{F1} \\
    \midrule
    Diacritics            & 1.00 & 1.00 & 1.00 \\
    Hallucination/Literal & 1.00 & 0.92 & 0.96 \\
    Loanword              & 0.91 & 0.82 & 0.86 \\
    No Error              & 0.93 & 0.97 & 0.95 \\
    Tense Shifting        & 0.77 & 0.84 & 0.80 \\
    \midrule[\heavyrulewidth]
    \textbf{Overall (macro-avg)} & \textbf{0.92} & \textbf{0.91} & \textbf{0.91} \\
    \bottomrule
  \end{tabular}
  \label{tab:judge}
\end{table}

\section*{Limitations \& Future Work}
Our study takes a first step toward Arabic-centric video-text retrieval, but richer domains, dialects and modalities remain wide open for exploration.

\vspace{0.5em}\noindent
\textbf{Generalization.} Our findings suggest that direct machine translation may enable language-agnostic benchmark replication without per-language retraining. Extending this beyond DiDeMo and MSA, across datasets, domains, and dialects, remains an open direction for future work.

\vspace{0.5em}\noindent
\textbf{Dataset Scope.}  DiDeMo-AR covers short clips (≤30\,s) captured in real-world conditions.  
Long-form videos such as movies, lectures, or sports broadcasts are out of scope. Future work could localize \textsc{MAD} corpus \cite{soldan2022mad} or the \textsc{LoVR} benchmark \cite{cai2025lovr}, for example, to MSA and dialects, giving researchers a benchmark for \emph{long-video retrieval}. 

\vspace{0.5em}\noindent
\textbf{Language Coverage.}  We focus on Modern Standard Arabic. Dialects, like: Egyptian, Gulf and Maghrebi, are still missing, yet they dominate social media videos \cite{GUELLIL2021497}. A fruitful extension is to repeat the framework for \emph{dialectal} captions.


\section*{Acknowledgments}
We are grateful to the KAUST Academy for its generous support, and especially to Prof.\ Sultan Albarakati and Prof.\ Naeemullah Khan for providing the resources and guidance that made this work possible.

\bibliography{custom}


\end{document}